\documentclass[runningheads]{llncs}

\usepackage{amssymb}
\usepackage{graphicx}

\usepackage{url}
\urldef{\mailsa}\path|{plangpra,lerman}@isi.edu|
\newcommand{\keywords}[1]{\par\addvspace\baselineskip
\noindent\keywordname\enspace\ignorespaces#1}

\long\def\comment#1{}

\newcommand{\coll}[1]{\textsf{#1}}
\newcommand{\set}[1]{\textsf{#1}}
\newcommand{\term}[1]{\texttt{#1}}

\newcommand{\eqref}[1]{Equation~\ref{#1}}
\newcommand{\figref}[1]{Figure~\ref{#1}}

\begin{document}

\mainmatter  % start of an individual contribution

\title{Constructing Folksonomies from User-specified Relations on Flickr}

\author{Anon Plangprasopchok\and Kristina Lerman}

\institute{USC Information Sciences Institute\\
4676 Admiralty Way, Marina del Rey, CA 90292\\
\mailsa\\
}

%\toctitle{Lecture Notes in Computer Science}
%\tocauthor{Authors' Instructions}
\maketitle

\begin{abstract}
Many social Web sites allow users to publish content and annotate with descriptive metadata. In addition to flat tags, some social Web sites have recently began to allow users to organize their content and metadata hierarchically. The social photosharing site Flickr, for example, allows users to group related photos in sets, and related sets in collections. The social bookmarking site Del.icio.us similarly lets users group related tags into bundles. Although the sites themselves don't impose any constraints on how these hierarchies are used, individuals generally use them to capture relationships between concepts, most commonly the broader/narrower relations.
Collective annotation of content with hierarchical relations may lead to an emergent classification system, called a folksonomy.
While some researchers have explored using tags as evidence for learning folksonomies, we believe that hierarchical relations described above offer a high-quality source of evidence for this task.

We propose a simple approach to aggregate shallow hierarchies created by many distinct Flickr users into a common folksonomy.
Our approach uses statistics to determine if a particular relation should be retained or discarded. The relations are then woven together into larger hierarchies. Although we have not carried out a detailed quantitative evaluation of the approach, it looks very promising since it generates very reasonable, non-trivial hierarchies.

\end{abstract}

\section{Introduction}
The subject of automatic taxonomy creation has attracted much attention from the academic community because of its close ties to important topics in philosophy, cognitive and computer sciences, and information technology. A taxonomy is a classification system that helps people organize their knowledge of the world hierarchically through broader-narrower (superclass-subclass) relations between concepts. One of the best known taxonomies is the Linnean classification of living organisms. There are alternative classification systems for organizing knowledge that do not rely exclusively on strict hierarchies. These include faceted classification schemes, which combine multiple taxonomies to represent objects, the various library classification schemes, such as the Dewey Decimal system, and Web directories, e.g., Yahoo directory and the Open Directory Project, which were created to categorize Web pages. Despite variations in structure, formal classification systems are distinguished by the fact that they use a \emph{controlled vocabulary} and are created by a small group of \emph{experts}. This means that formal classifications systems are often expensive to create and use, and it is difficult to keep them current in a fast-changing environment. Take, for example, Web directories. The first Web directory was created and is maintained by Yahoo, which hired a group of people to categorize Web pages. However, because Web changes at a rapid pace, with new pages added constantly and content of existing pages changing, it was difficult to keep the directory current. The Open Directory Project (ODP) attempted to mitigate some of these concerns by allowing a community of volunteers to edit a common Web directory. Although any user can register to become an editor, she still has to learn the structure and vocabulary and abide by the rules of the ODP.

As social Web sites, such as Flickr, Del.icio.us, and YouTube, become increasingly popular, massive amount of metadata about the content created by users is now available. The metadata comes in a variety of forms, including \emph{tags}, the freely-chosen keywords used to describe content, as well as links users create between content, metadata and other users. Although users annotate content for personal use, user-generated metadata can be used to discover relevant resources~\cite{delicious07::iiweb}, personalize search~\cite{flickr07}, and automatically generate         taxonomies~\cite{Mika07_OntoAreUs,SchmitzTagging06,ZhouBWY07}. Such a bottom-up taxonomy --- a \emph{folksonomy} --- has a number of advantages over formal top-down classification systems: (1) it is dynamic, evolving in time as community's needs and vocabulary change, (2) describes facets of data that are salient to users, and (3) has a level of detail that is meaningful to users. Similar to a formal classification system, an automatically generated folksonomy could be used for information management and discovery, as well as to annotate user-generated content in order to make it machine-readable.

The current approaches to automatic folksonomy creation combine tags created by large numbers of distinct individuals  by looking at statistics of their occurrence. It is possible that, because tags are flat, ambiguous and not expressive enough to annotate a large variety of content, social Web users began using inventions like colon ``:'' or slash ``/'' to combine several related keywords into a new tag. In many cases, the order of keywords glued by such separators have a meaning; for example, a preceding keyword is a superclass of the following keyword. Recognizing a demand, some social Web sites now allow users to specify some types of relations in addition to tags. Del.icio.us, for example, allows users to manually group related tags into \emph{bundles}, while, Flickr allows users to group related photos into \emph{sets} (similar to photo albums), and related sets into \emph{collections}.
Although the sites do not impose constraints on the semantics of relations expressed this way, we postulate that this type of metadata, both invented by users and available through social Web interfaces, is used to express ``broader/narrower'' relationships.
Users appear to categorize the content they create into shallow hierarchies, or taxonomies. We combine large numbers of such shallow hierarchies to infer a ``latent'' classification system, a folksonomy, that reflects the way individuals organize their knowledge. 

%Goal: we would like to learn hierarchies from user-provided partial hierarchies.

%problem with medata
Aggregating these relations into a folksonomy is not trivial because conflicts between users on certain relations may occur. One issue is noise, or the fact that some users will categorize content in a highly idiosyncratic manner. Another type of conflict is due to the individual difference in classification order. Suppose that user $A$ organizes her photos by creating a collection she calls \coll{travel}, and as part of this collection, a set called \set{china} for photos of her travels in China. Meanwhile user $B$ organizes her photos the other way round, by creating a collection \coll{china}, with constituent sets \set{travel}, \set{people}, etc.
Both categorizations are correct, since user $A$ might classify her photos in activity-oriented manner, as user $B$ in location-oriented manner. In addition to this, there are individual differences in the level of specificity: one user may organize photos first by country and then by city, while another organizes them by country, then subregion or state, and then city. Aggregating data from these users would potentially generate a ``shortcut'' from one concept to another., or multiple paths between concepts. Determining which
path is correct is a non-trivial issue. In addition to these challenges, there is also the familiar challenge of keyword ambiguity, where ``washington'' could mean the state or the city.

In this paper, we propose a simple framework for aggregating shallow individual hierarchies into a common folksonomy.\footnote{We call the learned concept structures folksonomies, even though they are not necessarily hierarchical.}
We use the shallow hierarchies created through the ``collection/set'' relations on Flickr. In this paper, we only resolve hierarchical relation conflicts due to noise, while leaving the issues of path selection and classification order for future work.
The contributions of these paper is as follows. First, we argue that partial hierarchies are a good source information for generating folksonomies. Second, we propose a simple statistical approach to resolve hierarchical relation conflicts in the aggregation process. Although we don't undertake a quantitative evaluation of the learner folksonomies, they appear to be very reasonable and detailed.

\section{Related work}
Many researchers have studied the problem of extracting ontological relations from text, \emph{e.g.}, \cite{Hearst92,Pasca04,Kozareva08}. These works exploit linguistic patterns to infer if two keywords are related under a certain relationship. For instance, they use ``such as'' (``vehicles, such as cars'') to learn hyponym relations. Cimiano \emph{et al.}~\cite{CimianoHS05} also applies linguistic patterns to extract object properties and then uses Formal Concept Analysis (FCA) to infer conceptual hierarchies. In FCA, a given object consists of a set of attributes and some attributes are common to a subset of objects. A concept `A' subsumes concept `B' if all objects in `B' (with some common attributes) are also in `A'. However, these approaches are not applicable to the metadata on the social Web such as tags, bundles and photo sets, which are ungrammatical and unstructured.

Recently, several papers proposed different approaches to construct conceptual hierarchies from tags collated from social Web sites.  Mika~\cite{Mika07_OntoAreUs} uses a graph-based approach to construct a network of related tags, projected from either a user-tag or object-tag association graphs. Although there is no evaluation of the induced broader/narrower relations, the work provides a good suggestion to infer them by using betweenness centrality and set theory.  Other works apply clustering techniques to keywords expressed in tags, and use their co-occurrence statistics to produce conceptual hierarchies~\cite{Brooks06,ZhouBWY07}. In a variation of the clustering approach, Heymann~\cite{Heymann06} uses graph centrality in the similarity graph of tags. In particular, the tag with the highest centrality would be more abstract than that with a  lower centrality; thus it should be merged to the hierarchy before the latter, to guarantee that more general node gets closer to the root node. Schmitz~\cite{SchmitzTagging06} has applied a statistical subsumption model~\cite{SandersonC99} to induce hierarchical relations of tags.

We believe that the previously mentioned works suffer from the ``popularity-generality'' problem that arises when using tags to induce a hierarchy. Specifically, a certain tag may be used more frequently not only because it is more general, but because it is more popular with users. In Flickr, for example, there are more photos tagged with \term{car} ($1,325,512$) than with \term{vehicle} ($71,498$).
% ``spider'' (166,113 photos) and ``arachnid'' (12,791 photos)
If we apply clustering approaches, \term{car} will be more general than \term{vehicle} since, the former is likely to have higher centrality than the latter. And if we apply statistical subsumption model, the former would be likely to subsume the latter since there is a higher chance that photos tagged with \term{vehicle} are also tagged with \term{car}. Of course, we believe that tag statistics are a good source of evidence for inducing hierarchies; however, tag statistics alone may not be adequate to distinguish between tag popularity and generality.

There is another line of research that focuses on exploiting partial hierarchies contributed by users. \emph{GiveALink} project~\cite{GiveALink06} collects bookmarks donated by users. Each bookmark is organized in a tree structure as folder and sub folders by an individual user. Based on tree structures, similarities between URLs are computed and used for URL recommendation and ranking. Although this project does not concentrate on conceptual hierarchy construction, it provides a good motivation to exploit explicit partial structures like folder and subfolder relations. Our approach is in the same spirit as \emph{GiveALink} --- we exploit collection and set relations contributed by users on a social Web site to construct conceptual hierarchies. We hypothesize that generality-popularity problem of keywords in collection-set relation space is less than that in tag space. Although people may use a keyword ``Washington'' far more than ``United States'' to name their collections and sets, not so many people would put their ``United States'' album into ``Washington'' super album, however.

% KL - need refs
Our approach is similar in spirit to ontology alignment, \emph{e.g.}, \cite{UdreaGM07}. However, unlike those works, which merge a small number of deep and detailed hierarchies, we merge large number of noisy, shallow hierarchies.

\section{Hierarchical structures on the social Web}
In addition to ``flat'' keywords or tags, some social Web sites have recently began to provide a feature that enables users to hierarchically organize content with broader/narrower relations. We believe that in the future many more social Web sites will allow their users to specify complex semantic relations, not only tags. We briefly describe how this feature is implemented on Flickr and del.icio.us.

%\comment{
\begin{figure}
\centering
\includegraphics[height=6.2cm]{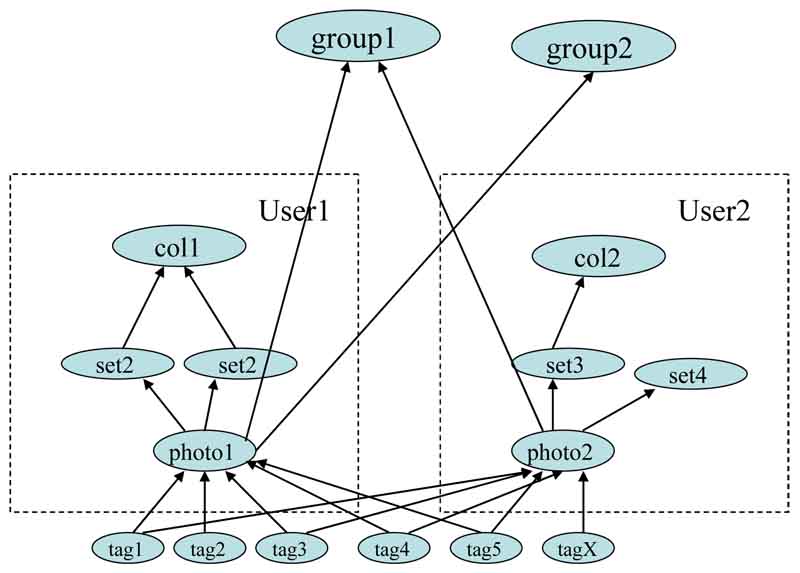}
\caption{A schematic view of how Flickr users organize their photos. A dot box cover an area controlled by a user, including content and metadata produced by the user. A photo can be annotated with a set of tags. Each photo can belong to one or more albums (sets); each album can be in a certain super-album (collection). A photo can also be submitted to a public group. An assignment of the photo to a group is independent to the set to which that photo belongs.}
\label{fig:flickr_hierarchical_scheme}
\end{figure}
%}

\begin{figure}
\center{
\begin{tabular}{c}
   \includegraphics[width=1.0\textwidth]{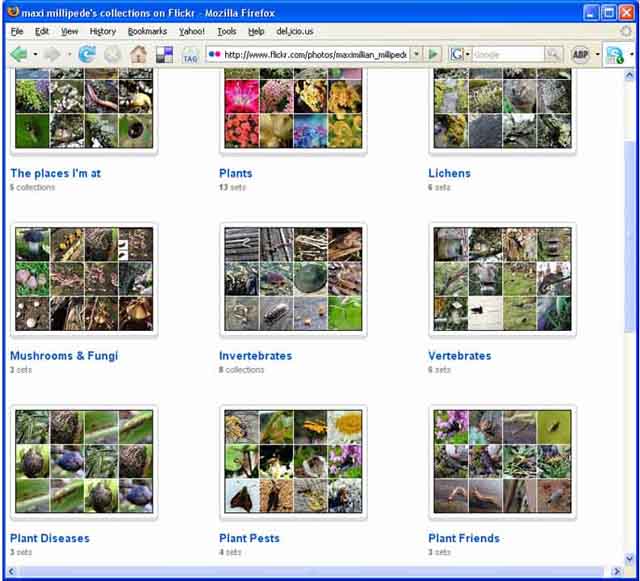} \\
  (a) \\
   \includegraphics[width=1.0\textwidth]{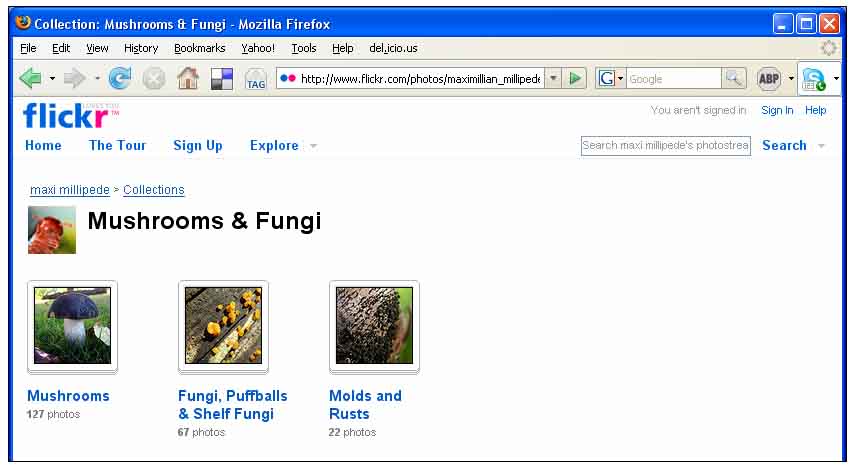} \\
(b)

\end{tabular}
}
  \caption{Personal hierarchy specified by a Flickr user. (a) Some of the collections created by this user and (b) the sets associated with one of the collections.}\label{fig:collections}
\end{figure}

The social photo-sharing site Flickr\footnote{http://www.flick.com} allows users to group their photos in album-like folders, named \emph{photo sets}. Users can also group their photo sets into ``super'' albums, called \emph{collections}. \footnote{The collection feature is limited to the paid ``pro'' users. Pro users can also create unlimited number of photo sets, while free membership limits a user to three sets.} \figref{fig:flickr_hierarchical_scheme} shows a schematic diagram of this organization. Both sets and collections are named by the owner of the images. Each photo can also be submitted to any of the thousands of special-interest groups Flickr users have created to share photos on a given topic.

Flickr does not enforce any specific rules about how to organize photos in sets and collections or how to name them. We found that most users group ``similar'' or ``related'' photos into the same set, and group related sets into the same collection, as shown in \figref{fig:collections}.
The name of the set generally subsumes all the photos within it, while the collection name is usually broad enough to cover all the sets within it.

On \emph{Del.icio.us},\footnote{http://del.icio.us} there is no explicit interface to group bookmarks into sets and collections as on \emph{Flickr}. Instead, users can group their tags into \emph{tag bundles}. This feature helps users to search and visualize tags as their number increases. Similar to sets and collections on Flickr, a user can assign an arbitrary name to a bundle. In general, the name of the bundle subsumes all associated tags.

\section{Aggregating relations from users}
In this paper, we only address the problem of inducing conceptual hierarchies from collection-set relations on Flickr. Such hierarchies contain concepts and broader/narrower relations.
% KL: what about ``related'' concepts?
We define $C^{i}$ a collection $i$  and $S^{ij}$ as a set $j$ of the $i$th collection.\footnote{A collection and its sets are specific to an individual user.} A collection or set names contain a series of terms: $<t_{1},...,t_{k}>^{C^{i}}$ is the name of $C^{i}$ and $<t_{1},...,t_{l}>^{S^{ij}}$ is the name of $S^{ij}$.

We assume that relations that a user specifies through collections and sets are broader-narrower relations. We denote that $C^{i}$ is broader that $S^{ij}$ as $C^{i} \rightarrow S^{ij}$.  These relations are also applicable to their constituent terms (relation delegation). In particular, if a user specifies the set $S^{ij}$ under the collection $C^{i}$ --- the former is narrower than the latter, and all the terms in $S^{ij}$ are also narrower than those of $C^{i}$. We also assume that each of those terms represents a concept in a conceptual hierarchy, and that the same terms used by the same or different users represent the same concept.\footnote{Although polysemy and synonymy do exist on Flickr, we ignore them for reasons of simplicity in this paper.}

\subsection{Approach}

From the problem definition above, we follow three main steps in aggregating relations: (1) term extraction and normalization; (2) relation conflict resolution; (3) concept prunning and linking. The first step is necessary because of variations in the names associated with the same concept, e.g., capitalization and punctuation. Thus, exact names are too sparse to be useful. Fortunately, we found that most of ``similar'' collections and sets share common terms. We use these instead of the full names and apply relation delegation as previously mentioned. The second step is necessary because of variations in the direction of relations among users. The last step prunes ``uninformative'' concepts and then links the rest into deeper hierarchies.

\subsubsection{Term extraction and normalization:} We tokenize collection and set names using simple heuristics. In addition to preposition words, users usually use characters such as `\&', `$<$', `$>$', `:', `/' to separate concepts. We, therefore, also tokenize names on these characters. We do not tokenize names on white spaces to avoid breaking up composite terms like \term{South Africa}. Non-alpha numeric characters and frequently-used non-informative words, \emph{e.g.}, ``me'' and ``myself'' are also removed. We then use Porter stemmer \cite{PorterStemmer} to normalize the remaining terms.

\subsubsection{Conflict resolution}: We assume that relation conflicts occur because of noise, because a minority of users specify relations opposite to the majority. For each relation, we simply consider how many users agree and disagree on it. Intuitively, concept $A$ subsumes concept $B$ if the number of users who agree on $A \rightarrow B$ is greater than the number who agree on $B \rightarrow A$, with some threshold. The formal formulation is as follows.

\comment{
\linebreak
\indent let $d_{x \rightarrow y}$ be the number of users who define $x \rightarrow y$\\
\indent and $d_{y \rightarrow x}$ be the number of users who define $y \rightarrow x$\\
\linebreak
\indent\indent 1. We define $x$ ``subsumes'' $y$ over all users (hard constraint) if: \\
\indent\indent\indent  $d_{x \rightarrow y} > 1$  and\\
\indent\indent\indent  $d_{y \rightarrow x} \leq 1$\\
\indent\indent 2. We define $x$ ``subsumes'' $y$ over all users (soft constraint) if:  \\
\indent\indent\indent  $d_{x \rightarrow y} > 1$  and\\
\indent\indent\indent  $d_{y \rightarrow x} \leq d_{x \rightarrow y}$ \\
}

 \begin{tabbing}
 xxx \= xxx \= xxx \= xxx \kill

\> let $d_{x \rightarrow y}$ be the number of users who define $x \rightarrow y$\\
\> and $d_{y \rightarrow x}$ be the number of users who define $y \rightarrow x$\\

\>\> 1. We define $x$ ``subsumes'' $y$ over all users (hard constraint) if: \\
\>\>\>  $d_{x \rightarrow y} > 1$  and\\
\>\>\>  $d_{y \rightarrow x} \leq 1$\\
\>\> 2. We define $x$ ``subsumes'' $y$ over all users (soft constraint) if:  \\
\>\>\>  $d_{x \rightarrow y} > 1$  and\\
\>\>\>  $d_{y \rightarrow x} \leq d_{x \rightarrow y}$ \\
 \end{tabbing}

In fact, the soft constraint (2) simply verifies that the number of users who agree on a relation is higher than the number of users who disagree. The hard constraint (1) is much more stringent, since it only allows at most one disagreement.

\subsubsection{Concept pruning and linking}: After the conflict resolution step, there are still some concepts which subsume too many other concepts, \emph{e.g.}, \term{all set}, \term{all rest}, \term{occasion}, and have few concepts subsume them. We feel that these ``uninformative'' concepts seem to be too broad to be useful. From our informal analysis, we postulate that a number of parent and child concepts can be used to determine if a concept is uninformative. The formulation for this heuristic is provided as follows.

\comment{
\linebreak
\indent let $din_{x}$ be the number of parent concepts of the concept $x$ (in-degree)\\
\indent and $dout_{x}$ be the number of child concepts of the concept $x$ (out-degree)\\
\indent\indent We define ratio $R_{x^{oi}}$ as $dout_{x}/din_{x}$. \\}

 \begin{tabbing}
 xxx \= xxx \= xxx \= xxx \kill

 \> let $din_{x}$ be the number of parent concepts of the concept $x$ (in-degree)\\
\> and $dout_{x}$ be the number of child concepts of the concept $x$ (out-degree)\\
\>\> We define ratio $R_{x^{oi}}$ as $dout_{x}/din_{x}$. \end{tabbing}

In particular, we found that $R_{x^{oi}}$, can indicate if $x$ is uninformative: the higher the ratio, the more uninformative the concept $x$ is. In many concepts, they have no parent concepts and divided-by-zero can occur. To avoid such, we smooth both $din_{x}$ and $dout_{x}$ with a very small number relative to a number of all concepts. After pruning uninformative concepts, concepts are then linked together through their subsumption relations.
% KL: subsumption relations.?
%relations under the assumption we mentioned so far.

\subsection{Case Study}
\label{sec:study}
%- Data collection process
For our study, we gathered data about collection/set relations created by a set of Flickr users, identified by their ids. To gather user ids, we used the Flickr API to retrieve the names of members of seventeen public groups devoted to wildlife, specifically insect, photography. 
We then used a Web page scraping tool to retrieve the names of collections and sets created by these users. Although a small fraction of users created multi-level hierarchies, we only retrieved names associated with the top two levels. Of the $39,922$ users in our set, $21,792$ created at least one collection.
%i'm not sure about $36,632$ users (I only have 6967+22959 = 29926)

%- some statistics on collected data
After processing data, we obtain $6,871$ and $7,196$ out of $213,104$ relations using hard and soft constraint respectively. The number of concepts is reduced from $94,499$ to $3,239$ and $3,244$ concepts for hard and soft constraint respectively. Top 200 concepts with high $R_{x^{oi}}$ are discarded.

\begin{figure}[tbhp]
  \includegraphics[width=1.2\textwidth]{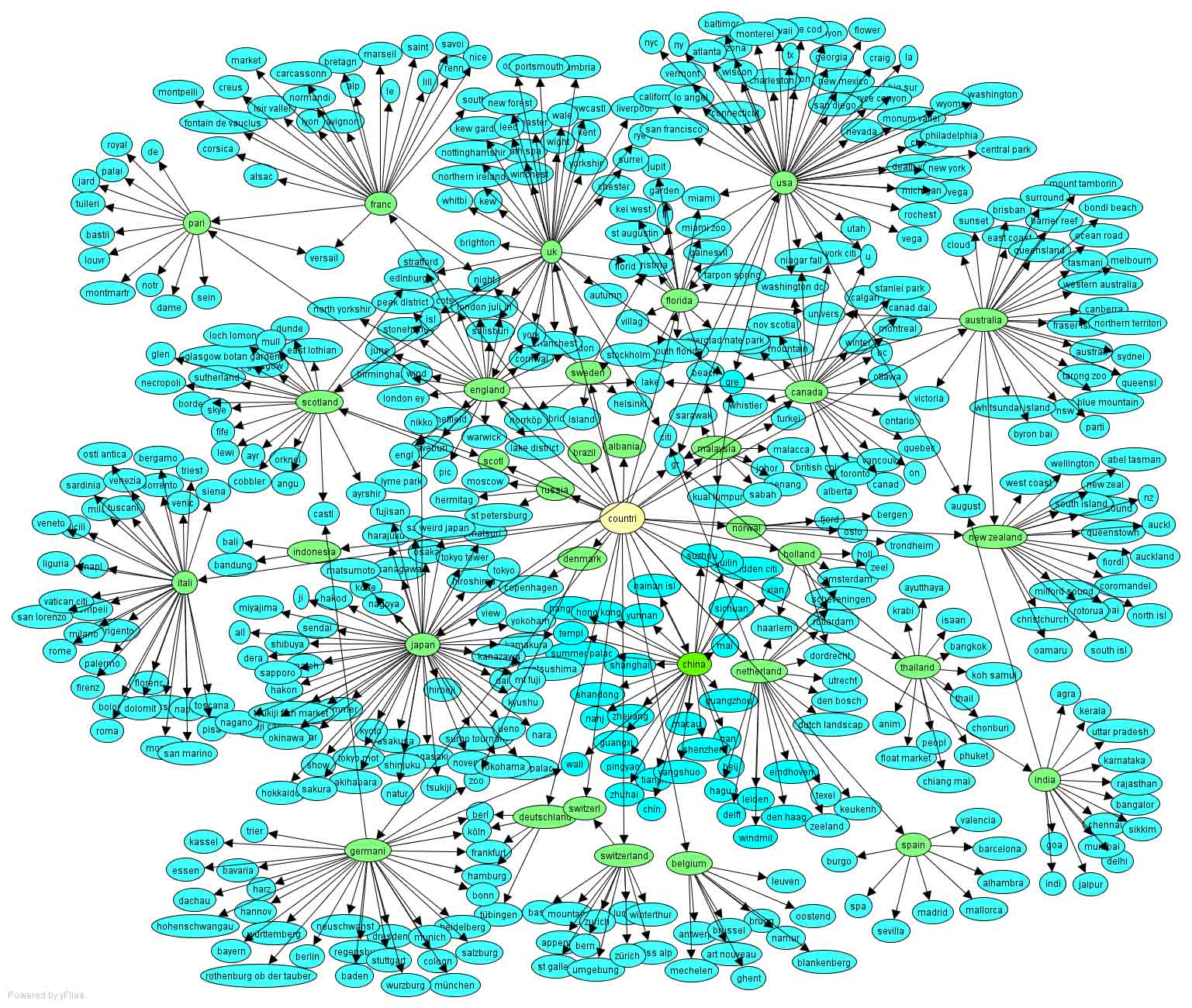}\\
  \caption{Folksonomy associated with the concept \term{country}, showing its broader and  narrower concepts. We colored the graph to aid visualization. The starting concept is in yellow, its parent concepts (where applicable) are in pink, while the direct descendants are in green. The rest of the descendants are in blue.}\label{fig:countri}
\end{figure}

The resulting graph of interlinked concepts is quite complex. To simplify browsing, we extract subgraphs associated with a concept. Starting with a given concept, we get its parents (broader concepts), and then follow the outgoing links on the graph to get the children (narrower concepts) and the children's children, etc. We illustrate the results with some sample graphs. We colored the graphs to aid visualization. The starting concept is in yellow, its parent concepts (where applicable) are in pink, while the direct descendants are in green. The rest of the descendants are in blue. The graph in \figref{fig:countri} shows the concept graph for the (stemmed) \term{country}.   It has $32$ children (in green), including \term{france}, \term{china}, \term{india}, \term{uk}, etc. Of the $32$ children only two, \term{florida} and \term{paris}, are not proper countries. The concepts in blue are the children of the individual countries. In general, these automatically discovered concepts are quite good. For example, \term{russia} has narrower concepts \term{moscow}, \term{st petersburg} and \term{hermitage}, while \term{england} has \term{warwick}, \term{stonehenge}, and \term{lake district}, among others.

\begin{figure}[tb]
\begin{tabular}{c}
  \includegraphics[width=0.7\textwidth]{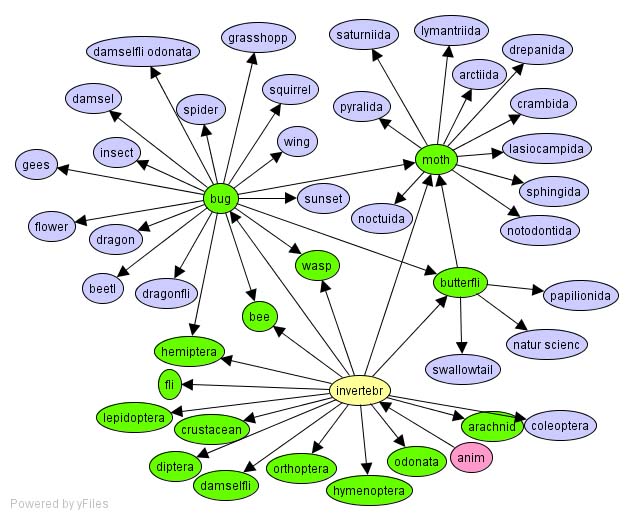} \\
  (a) \\
  \includegraphics[width=0.5\textwidth]{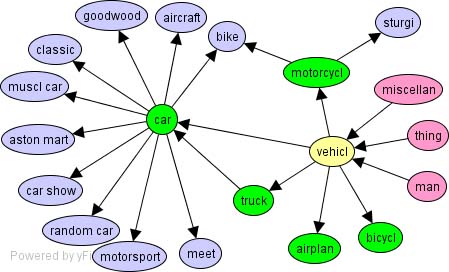}\\
(b)
\end{tabular}
  \caption{Folksonomy associated with the concept (a) \term{invertebrate} and (b) \term{automobile} showing their  broader and  narrower concepts.}\label{fig:invertebr}
\end{figure}

While geographical names provide a common vocabulary for labeling and organizing travel photographs, there is sufficient vocabulary agreement to induce concept graphs in many other domains. \figref{fig:invertebr} shows the graphs associated with (a) \term{invertebrate} and (b) \term{vehicl}. The parent concept  (in pink) of \term{invertebrate} in the first graph is \term{animal}. This graph reflects the bias in our data --- we collected relations from users who have posted to Flickr groups related to insect photography.  These users had diverse enough interests though, and have apparently also expressed knowledge about the various modes of transportation. The child concepts are \term{vehicle} are \term{car}, \term{bike}, \term{truck}, \term{bicycle}, \term{motorcycle}, and \term{airplan}. Not all the subsumption relations make sense, but overall, they are quite useful.

\begin{figure}[tb]
  \includegraphics[width=0.8\textwidth]{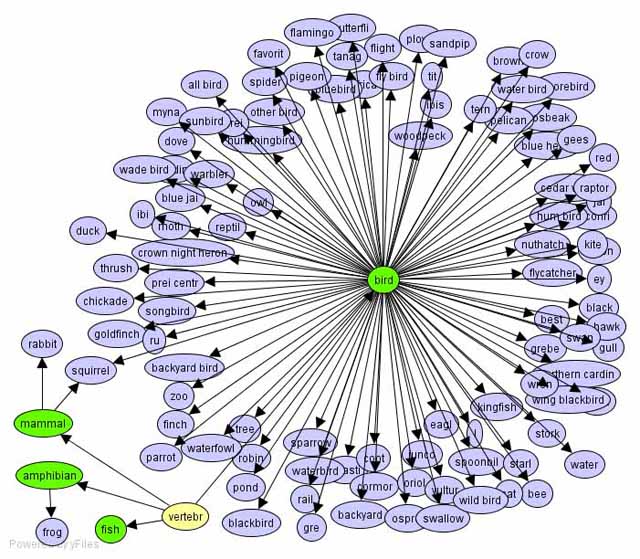}\\
  \caption{Folksonomy associated with the concept \term{invertebrate}, showing its broader and narrower concepts.}\label{fig:vertebr}
\end{figure}

\begin{figure}[tb]
  \includegraphics[width=0.8\textwidth]{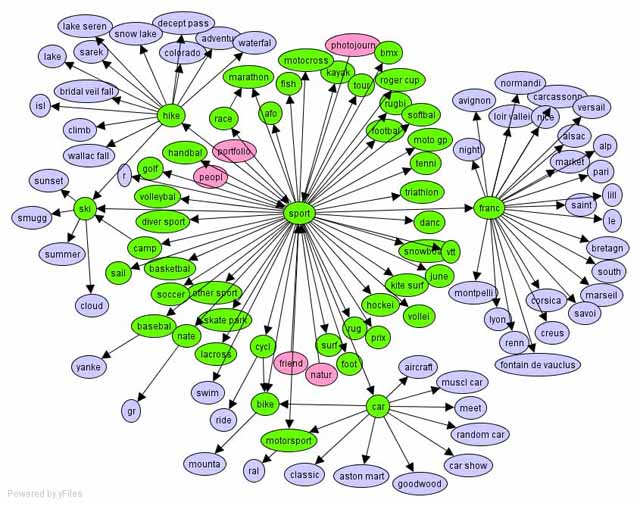}\\
  \caption{Concept graph associated with the concept \term{invertebrate}, showing its broader and related narrower concepts.}\label{fig:sport}
\end{figure}

We present two more concept graphs to illustrate our method's ability to discover many relevant subcategories. \figref{fig:vertebr} shows the graph associated with the concept \term{vertebrate}, which includes \term{bird} and many concepts corresponding to specific types of birds. Similarly, \term{sport} graph shows many specific types of sports. Our algorithm associated \term{france} with \term{sport}, maybe because of the popular `tour de france'. However, all other discovered subsumption relations are correct.

We also apply the term-based (as opposed to relation-based) subsumption approach~\cite{SchmitzTagging06} on the dataset we collected. In particular, we tokenize and normalize collection and set names as the same way we did before. For each set, we aggregate its terms and collection terms together as a document (a bag of concepts). We hypothesize that terms used in collection names are more prevalent (and thus have high centrality) --- and subsume --- terms from set names. We then use subsumption approach with the threshold specified in \cite{SchmitzTagging06}. The hierarchies produced by this approach are much sparser and contain fewer informative concepts than the folksonomy learned by our approach. We also tried to relax subsumption threshold in steps from $0.8$ to $0.55$; yet many informative concepts and relations were still discarded.

% I'm not sure about the following statement -- this is just from my justification but this challengs existing statistical models based only from word co-occurrences
One reason why subsumption approach does not work very well in this context is that a certain concept usually relates to many other concepts. Thus, it is very likely that a number of co-occurrences of a given concept pair is very low, compared to that of individual one. Consequently, a chance that one concept ``subsumes'' another one is very low. From our dataset, we found that a relation between \term{china} and \term{countri} is not induced by the subsumption approach. In particular, a number of their co-occurrences is just $6$ compared to their frequency $596$ and $256$ respectively, and consequently neither is judged to subsume the other. In our approach, we instead consider explicit relations of concepts, which will not suffer from this issue.

% Claim: will publish this in future work..
% We would think that subsumption approach would work well when concepts are only lying on a certain conceptual path i.e. there is only one facet for one concept.  If we only have a -> b -> c, subsumption will work properly. But if we also have {d, e, f} -> c, subsumption will not work since p(c) is splitted into p(c,b), p(c,d), p(c,e), p(c,f). Thus b will no longer subsume c since p(b,c) is too small. Let's think about {travel, countri} -> {china} case!!!

\section{Conclusion}
The social Web sites allow users to contribute content and also provide tools to help them manage content by annotating it with descriptive tags, and more recently, with semantic relations. By making large amount of such metadata available, social Web sites enable researchers to empirically study how humans organize knowledge, and also to learn a common classification system, a folksonomy, from the data. This paper describes a statistical approach to aggregating large number of simple broader/narrower relations specified by different users into a common, deeper folksonomy. The broader/narrow relations we used for the study were expressed through the shallow hierarchies of photo sets and collections created by Flickr users to manage their photos. Our approach is general, and can be applied to other systems that allow users to specify relations: e.g., the social bookmarking site Del.icio.us allows users to group related tags into tag bundles.

Our long-term goal is to learn the structure of collective knowledge from the evidence provided by many users~\cite{CKemp05}. We believe that the simple relations described above are more informative than tags alone for learning how people classify things. Although we have not quantitatively compared the folksonomy learned by our approach to existing classification systems, qualitative evaluation indicates that our baseline method already yields good quality folksonomies. There is still much room for improvement. In the future, we plan to separate ``broader/narrower'' from ``related-to'' relations. We also need to more systematically handle the challenges of different users using a different classification order and different level of specificity in the relations they specify. We would also like to combine relations with tag statistics to disambiguate concepts. We would also like to perform a systematic evaluation of the learned folksonomies, e.g., by comparing learned structures to ODP's dmoz, the open Web directory.

%- This approach can be applied to delicious and other systems../ plan to test on delicious
%        + utilize less data but having high information (since user explicitly express relations)
%        + separate boader/narrower from just-related relations
%        + combine to tag statistics / neighbor concepts to disambiguate terms
%- Evaluation (dmoz comparison and task-based evaluation)
%- Long term goal : cognitive perspective? -> challenging issue is how to learn actual structures of things from evidences provided by each user who has different (1) classification order and (2) level of specificity

\subsubsection*{Acknowledgements}
We would like to thank Fetch Technologies for providing us with a Web page scraping tool. We also appreciate yWorks for providing yEd freely available for visualizing concept relations.

\bibliographystyle{splncs}
\bibliography{hierfolk}
\end{document}